\documentclass[nohyperref]{article}

\usepackage{microtype}
\usepackage{graphicx}
\usepackage{booktabs}
\usepackage{arydshln}
\usepackage{tabu}
\usepackage{hyperref}
\usepackage{xspace}
\usepackage{multirow}
\usepackage{caption}
\usepackage{subcaption}

\usepackage{enumitem}
\usepackage{bbm}

\usepackage[accepted]{icml2023}

\usepackage{amsmath}
\usepackage{amssymb}
\usepackage{mathtools}
\usepackage{amsthm}
\usepackage{latexsym}
\usepackage{pifont}

\usepackage[capitalize,noabbrev]{cleveref}

\theoremstyle{plain}

\theoremstyle{definition}

\theoremstyle{remark}

\usepackage[capitalize,noabbrev]{cleveref}
\crefname{section}{\S}{\S\S}
\Crefname{section}{\S}{\S\S}
\crefname{table}{Table}{Tables}
\crefname{figure}{Figure}{Figures}
\crefname{algorithm}{Algorithm}{}
\crefname{equation}{eq.}{}
\crefname{appendix}{Appendix}{}
\crefformat{section}{\S#2#1#3}
\usepackage{float}

\usepackage[textsize=tiny]{todonotes}

\begin{document}

\twocolumn[
%\icmltitle{Large Language Models Are Sensitive to Irrelevant Context}
\icmltitle{Large Language Models Can Be Easily Distracted by Irrelevant Context}

% It is OKAY to include author information, even for blind
% submissions: the style file will automatically remove it for you
% unless you've provided the [accepted] option to the icml2022
% package.

% List of affiliations: The first argument should be a (short)
% identifier you will use later to specify author affiliations
% Academic affiliations should list Department, University, City, Region, Country
% Industry affiliations should list Company, City, Region, Country

% You can specify symbols, otherwise they are numbered in order.
% Ideally, you should not use this facility. Affiliations will be numbered
% in order of appearance and this is the preferred way.
\icmlsetsymbol{equal}{*}

\begin{icmlauthorlist}
% Freda: temporary author list, open to debate
\icmlauthor{Freda Shi}{googleresearch,ttic,equal}
\icmlauthor{Xinyun Chen}{googleresearch,equal}
\icmlauthor{Kanishka Misra}{googleresearch,purdue}
\icmlauthor{Nathan Scales}{googleresearch}
\icmlauthor{David Dohan}{googleresearch}
\icmlauthor{Ed Chi}{googleresearch}

\icmlauthor{Nathanael Sch\"arli}{googleresearch}
\icmlauthor{Denny Zhou}{googleresearch}
\end{icmlauthorlist}

\icmlaffiliation{googleresearch}{Google DeepMind}
\icmlaffiliation{ttic}{Toyota Technological Institute at Chicago}
\icmlaffiliation{purdue}{Purdue University}

\icmlcorrespondingauthor{Freda Shi}{freda@ttic.edu}
\icmlcorrespondingauthor{Xinyun Chen}{xinyunchen@google.com}
\icmlcorrespondingauthor{Denny Zhou}{dennyzhou@google.com}

% You may provide any keywords that you
% find helpful for describing your paper; these are used to populate
% the "keywords" metadata in the PDF but will not be shown in the document
%\icmlkeywords{Machine Learning, ICML}

\vskip 0.3in
]

% this must go after the closing bracket ] following \twocolumn[ ...

% This command actually creates the footnote in the first column
% listing the affiliations and the copyright notice.
% The command takes one argument, which is text to display at the start of the footnote.
% The \icmlEqualContribution command is standard text for equal contribution.
% Remove it (just {}) if you do not need this facility.

\printAffiliationsAndNotice{Work done while FS and KM are student researchers at Google DeepMind. \icmlEqualContribution}
% leave blank if no need to mention equal contribution
%\printAffiliationsAndNotice{\icmlEqualContribution} % otherwise use the standard text.

\newcommand{\chainofthought}{\textsc{CoT}\xspace}
\newcommand{\leasttomost}{\textsc{LtM}\xspace}
\newcommand{\program}{\textsc{Program}\xspace}
\newcommand{\noisygsm}{\textsc{GSM-IC}\xspace}
\newcommand{\selfconsistency}{SC\xspace}
\newcommand{\zeroshotcot}{\textsc{0-CoT}\xspace}
\newcommand{\nlinstruction}{\textsc{NL-Instruction}}
\newcommand{\interalia}[1]{\citep[\textit{inter alia}]{#1}}

\newcommand{\eat}[1]{}

% author comments
\newcommand{\freda}[1]{\textcolor{orange}{[Freda: #1]}}
\newcommand{\xinyun}[1]{\textcolor{blue}{[Xinyun: #1]}}
\newcommand{\km}[1]{\textcolor{teal}{[Kanishka: #1]}}

% anonymity commands
\newcommand{\datasetlink}{\texttt{AnonymousLink}}

% special chars
\newcommand{\cmark}{\ding{51}}
\newcommand{\xmark}{\ding{55}}

% writing commands
\makeatletter
\def\adl@drawiv#1#2#3{%
        \hskip.5\tabcolsep
        \xleaders#3{#2.5\@tempdimb #1{1}#2.5\@tempdimb}%
                #2\z@ plus1fil minus1fil\relax
        \hskip.5\tabcolsep}
\newcommand{\cdashlinelr}[1]{%
  \noalign{\vskip\aboverulesep
           \global\let\@dashdrawstore\adl@draw
           \global\let\adl@draw\adl@drawiv}
  \cdashline{#1}
  \noalign{\global\let\adl@draw\@dashdrawstore
           \vskip\belowrulesep}}
\makeatother
\newif\ifsubmission
\begin{abstract}
Large language models have achieved impressive performance on various natural language processing tasks. 
However, so far they have been evaluated primarily on benchmarks where all information in the input context is relevant for solving the task.
In this work, we investigate the \emph{distractibility} of large language models, i.e., how the model problem-solving accuracy can be influenced by irrelevant context.
In particular, we introduce Grade-School Math with Irrelevant Context (\noisygsm), an arithmetic reasoning dataset with irrelevant information in the problem description.
We use this benchmark to measure the distractibility of cutting-edge prompting techniques for large language models, and find that the model performance is dramatically decreased when irrelevant information is included.
%the model is easily distracted by irrelevant information.
We also identify several approaches for mitigating this deficiency, such as decoding with self-consistency and adding to the prompt an instruction that tells the language model to ignore the irrelevant information.\footnote{Dataset is available at \url{https://github.com/google-research-datasets/GSM-IC}.}

%Large language models have achieved impressive performance on various natural language processing tasks. 
%However, so far they have been evaluated primarily on benchmarks where all information in the input context is relevant for solving the task.
%\xinyun{TODO: we introduce distractibility as a new metric for evaluating language models.}
%In this work, we introduce Grade-School Math with Irrelevant Context (\noisygsm), an arithmetic reasoning dataset with irrelevant information in the problem description.
%We find that the state-of-the-art language models are sensitive to the irrelevant context in the prompt, whereas decoding with self-consistency significantly boosts the performance.
%In addition, we find that when adding to the prompt an instruction that explicitly requires the language models to ignore irrelevant information, the model impressively follows the instruction to some degree and achieves higher accuracies.
%The \noisygsm dataset can be found at \datasetlink. 
% {\color{red} (Denny: we should replace all "noise" stuff with "irrelevant contexts" )}
\end{abstract}
\section{Introduction}
%Recent work has demonstrated that 
Prompting large language models performs decently well in a variety of domains \interalia{brown2020language, chowdhery2022palm}. %such as arithmetic reasoning \citep{cobbe2021training}, commonsense reasoning \citep{roemmele2011choice} and question answering \citep{yang2018hotpotqa}.
However, for most of theses evaluation benchmarks, all the information provided in the problem description is relevant to the problem solution, as the problems in exams.
This is different from real-world situations, where problems usually come with several pieces of contextually related information,  which may or may not be relevant to the problems that we want to solve. We have to identify what information is actually necessary during solving those problems. Studies in psychology have shown that irrelevant information
may significantly decrease some children and even adults problem-solving accuracy \interalia{hoyer1979effects, pasolunghi1999working, marzocchi2002disturbing}. %and part of the solution requires identifying what information is actually relevant.
%As an example, consider the original problem in \cref{fig:teaser}. %People who know Jeanne would have various additional information about her, such as her age, her height, and the commuting distance of her neighbor.

In this work, we study the \emph{distractibility} of large language models for various prompting techniques; i.e., how is large language model prompting affected by irrelevant context, and what strategies can be used to improve performance?
To measure distractibility, we construct the \noisygsm dataset, a grade-school math problem dataset derived from GSM8K \citep{cobbe2021training} and introduce two different metrics.
In contrast to prior work that derives benchmark variations by substituting sentences of the base problems with variations \interalia{patel2021nlp,kumar2021adversarial}, we keep the base problem description and add to it one irrelevant sentence, while making sure that it does not affect the solution of the problem (\cref{fig:teaser}). 
%All \noisygsm examples are based on ``easy'' GSM8K problems that can be $100\%$ solved by multiple prompting techniques with greedy decoding.
\begin{table}[t]
    \centering
    \begin{tabular}{p{0.45\textwidth}}
    \toprule
    \textbf{Original Problem} \\
    Jessica is six years older than Claire. In two years, Claire will be 20 years old. How old is Jessica now? \\[10pt]
    \normalsize \textbf{Modified Problem} \\
    Jessica is six years older than Claire. In two years, Claire will be 20 years old. \textcolor{red}{\underline{\it Twenty years ago, the age of} \underline{\it Claire’s father is 3 times of Jessica’s age.}} How old is Jessica now? \\[10pt]
    \normalsize \textbf{Standard Answer} 24 \\
    \bottomrule
    \end{tabular}
    
    \iffalse 
    \includegraphics[width=0.45\textwidth]{rawfigs/teaser.pdf} \vspace{-10pt}
    \fi 
    \caption{An example problem from \noisygsm. 
    An irrelevant sentence (\underline{\textit{italic and underlined}}) that does not affect the standard answer is added immediately before the question.}
    \label{fig:teaser}
\end{table}

We use Codex (\texttt{code-davinci-002}) and GPT-3.5 ( \texttt{text-davinci-003}) in the GPT3 model family to evaluate state-of-the-art prompting techniques on \noisygsm,\footnote{\url{http://openai.com/api/}} including chain-of-thought prompting \citep[\chainofthought;][]{wei2022chain}, zero-shot chain-of-thought prompting \citep[0-\chainofthought;][]{kojima2022large}, least-to-most-prompting \citep[\leasttomost;][]{zhou2022least}, and prompting with programs \citep[\program;][]{chowdhery2022palm}.%,gao2022pal,chen2022program}.  
 We find that their performance on \noisygsm greatly decreases compared to the original GSM8K (without irrelevant context).
We then investigate several approaches to mitigate this weakness, including self-consistency \citep{wang2022self} and adding irrelevant information to the exemplars in the prompt. In addition to demonstrating how to handle irrelevant information via exemplars, we also investigate the usage of task-specific instructions \citep{wei2021finetuned,sanh2021multitask,ouyang2022training, suzgun2022challenging, chung2022scaling}, where we prepend an instruction sentence \textit{``feel free to ignore irrelevant information in the problem description''} to the exemplars. We summarize our key findings below:
\begin{enumerate}[leftmargin=*,itemsep=0pt,topsep=0pt,parsep=2pt,partopsep=0pt]
\item All investigated prompting techniques are sensitive to irrelevant information in the problem description. In particular, among the original problems that can be solved by baseline prompts with greedy decoding, no more than $18\%$ of them can be consistently solved for all types of irrelevant information, showing that the large language model is easily distracted and produces inconsistent predictions when adding a small amount of irrelevant information to the problem description.
\item Self-consistency improves the performance of all prompting techniques on \noisygsm. 
In particular, the recall rate of the correct answer for \noisygsm is as high as 99.7\% with 20 samples per problem, i.e., at least one of the 20 solutions result in the correct final answer, which means that using multiple samples allows the model to almost always retrieve the correct answer.
\item Adding irrelevant information to the exemplars shown in the prompt consistently boosts the performance, and the same holds for adding an instruction to ignore irrelevant context.
This suggests that language models are---to some extent---able to learn to ignore irrelevant information by following examples or instructions.
\item We identify different factors of the irrelevant information that affect the model's sensitivity to irrelevant context. Our breakdown analysis shows that varying the numbers in the irrelevant information does not notably change the model performance, while the degree of lexical overlap with the original problem description matters.
\end{enumerate}
Filtering out irrelevant information is essential for handling real-world tasks. Our evaluation indicates that despite the strong performance on challenging reasoning problems, state-of-the-art language models still have fundamental weaknesses in context understanding and identifying the relevant information from the input. 
Our findings suggest that in order to gain a more holistic understanding of the reasoning capability of language models, future work should also consider the model sensitivity to irrelevant context, in addition to solving more challenging problems.
\section{Related Work}
% \freda{needs update.}
%\paragraph{Large language models and prompting.} 

\noindent \textbf{Few-shot prompting.} 
%Recent work \interalia{brown2020language,raffel2020exploring,ouyang2022training} demonstrates that large language models achieve impressive performance on various tasks via few-shot prompting, including arithmetic reasoning \citep{cobbe2021training,patel2021nlp}, common-sense reasoning \citep{roemmele2011choice,talmor2019commonsenseqa,geva2021did} and question answering \citep{yang2018hotpotqa}.
%Their performance 
Few-shot prompting \interalia{brown2020language, chowdhery2022palm} has been significantly boosted with various techniques, including generating intermediate steps \interalia{ling-etal-2017-program, cobbe2021training, nye2021show, wei2022chain, suzgun2022challenging, shi2022language}, problem decomposition \interalia{zhou2022least, drozdov2022compositional, dohan2022language, khot2022decomposed, press2022measuring}, generating programs \interalia{austin2021program, chowdhery2022palm,gao2022pal,chen2022program}, marginalizing intermediate steps that share the same result \citep{wang2022self,shi2022natural}, and ensemble \citep{wang2022rationale, drozdov2022compositional}.
In addition, \citet{kojima2022large} demonstrate that appropriate hint in prompts also leads to decent performance, even without any exemplar.
In this work, we examine these cutting-edge prompting techniques \citep{wei2022chain,zhou2022least,kojima2022large,wang2022self} on our benchmark, and demonstrate that they are sensitive to irrelevant input context.

\noindent \textbf{Natural language benchmarks with input perturbations.}
There has been a long line of work on adding input perturbations for natural language tasks, including model-agnostic input transformations \interalia{liang2022holistic,ravichander2022condaqa} and adversarial example generation against individual models \citep{jia2017adversarial,shi2018learning,morris2020textattack,wang2021adversarial}.
In particular, prior work has constructed arithmetic reasoning benchmarks through paraphrasing or rewriting sentences in the base problems from clean datasets \citep{patel2021nlp,kumar2021adversarial}. 
Meanwhile, \citet{liang2022holistic} evaluate various large language models under several metrics, including accuracy, robustness, fairness, etc. Specifically, the input transformations in their robustness evaluation include semantics-preserving and semantics-altering perturbations, such as injecting typos and modifying sentences to change the ground-truth classification labels.
In contrast the above work where the meaning of problem descriptions may be changed with perturbations, we keep all sentences in the original problem description, and introduce an irrelevant sentence that is ensured not to affect the standard answer.

\noindent\textbf{Natural language benchmarks with irrelevant input context.} 
\citet{jia2017adversarial} have shown that neural question answering systems are largely affected by adversarial distracting sentences, whereas follow up work \citep{khashabi-etal-2017-learning,ni-etal-2019-learning} proposes learning strategies that mitigate the problem. 
Similar issues have been found for general-purpose pretrained language models, on the tasks of factual reasoning \citep{kassner-schutze-2020-negated, pandia-ettinger-2021-sorting, misra2022comps, li2022large}, code generation \citep{jones2022capturing}, and syntactic generalization \citep{chaves-richter-2021-look}. 
In particular, \citet{li2022large} evaluated T5 \citep{raffel2020exploring} and PaLM \citep{chowdhery2022palm} with few-shot prompts, and proposed knowledge-aware finetuning that finetunes the model on problems with counterfactual and irrelevant context, which strengthens the model robustness to noisy context. 
In our evaluation, we show that without training or finetuning, adding irrelevant context into demonstrations in the prompt also mitigates the distractibility of the underlying language model and significantly improves the model performance on our GSM-IC benchmark.

\iffalse
Distractor related work:
- Semantic distraction: 
    Jia and Liang (2017)
    Learning important information 
        https://aclanthology.org/K17-1010/
        https://aclanthology.org/N19-1030/
    Distracts BERT for probing with negation and priming.    
        https://aclanthology.org/2020.acl-main.698.pdf 
    Probably move the line of Pandia and Ettinger (2021) and related work here. 
- Syntactic distraction: 
    Easily distracted from generalizing syntactically https://aclanthology.org/2021.scil-1.3.pdf
- Cite a few multi-choice cloze tasks, as distractor choice is often used there. Some pointers can be found from Kevin's paper: https://aclanthology.org/2020.bea-1.10.pdf 
\fi

There exist some logical reasoning benchmarks that contain irrelevant content in task descriptions \citep[\textit{inter alia}]{weston2015towards, sinha-etal-2019-clutrr, clark2021transformers,han2022folio,tafjord2020proofwriter}. However, previous work largely focuses on designing models that require extra training, and prompting alone still hardly achieves the same level of performance as finetuned models for these tasks \citep{han2022folio,creswell2022selection}. In our work, we focus on arithmetic reasoning, where prompting techniques have achieved the state-of-the-art results, e.g., on GSM8K, while we show that adding a single irrelevant sentence into the problem description significantly degrades the performance.

\noindent\textbf{Prompting with noisy ground truth.} 
A line of work studies the model performance with incorrect prompting exemplars, i.e., the example problems are paired with wrong answers \citep{min2022rethinking,kim2022ground}. 
In addition, prior work has investigated the model sensitivity to other parts of the prompt, such as instruction tuning with misleading and irrelevant instructions \citep{webson2021prompt} and wrong reasoning steps in the examples \citep{madaan2022text,wang2022towards}. 
In particular, \citet{madaan2022text} conclude that the correctness of numbers and equations in chain-of-thought prompts does not play a key role in model performance, but using wrong entities and removing either equations or text explanation in the reasoning steps drastically hamper the performance.
% In particular, \citet{madaan2022text} study the role of different factors in chain-of-thought prompts for large language models; for example, on GSM8K \citep{cobbe2021training}, they evaluate model performance on variants of prompts that contain wrong equations or wrong entities in reasoning steps, replace numbers with Greek symbols, replace numbers with out-of-distribution numbers, etc. 
% They conclude that the correctness of numbers and equations in chain-of-thought prompts does not play a key role in model performance, but using wrong entities and removing either equations or text explanation in the reasoning steps drastically hamper the performance.
Different from this line of work, we always include correct answers to example problems in the prompt, and ensure that the irrelevant context added to the problem description does not change the ground truth answer. 
We show that the model performance significantly drops when presented with irrelevant context in problem descriptions, and different distributions of numbers and entities in the irrelevant context also lead to different levels of performance degradation.

\vspace{-1em}
\section{The \noisygsm Dataset}
\begin{table}[t]
    \centering \small
    \begin{tabular}{ccccc}
    \toprule
       &  \bf \chainofthought &  \bf \leasttomost & \bf \program & \bf 0-\chainofthought \\
    \cdashlinelr{1-5} 
       &  95.0 & 94.0 & 83.0 & 44.0 \\
     + SC & 96.0 & 99.0 & 91.0 & 76.0 \\
    \bottomrule
    \end{tabular}
    \caption{ Accuracy ($\times 100$) on the base 100-example dataset using \texttt{code-davinci-002}. See \cref{tab:main-results} for results with \texttt{text-davinci-003}.}
    \label{tab:base-accu}
\end{table}

In this section, we introduce the creation process of the \noisygsm dataset (\cref{sec:dataset}) and the evaluation metrics (\cref{sec:metrics}).
\subsection{Dataset Creation}
\label{sec:dataset}
We randomly choose 1,000 problems from the GSM8K training set as a development set.
To construct our base dataset, we then choose 100 problems from this development set that can be correctly solved by at least one of the prompting techniques mentioned in this paper;\footnote{
We do not generate new examples or perform analysis on the test set to avoid potential tuning-on-test-set issues.
}
that is, our base dataset is an ``easy'' subset of GSM8K (\cref{tab:base-accu}).
Each base problem requires two to seven reasoning steps to solve.\footnote{The number of reasoning steps of a problem is given by the number of sentences in its standard answer \citep{cobbe2021training}.} Among the 100 base problems, 60 of them can be solved with two reasoning steps. The full dataset statistics can be found in \cref{sec:appendix-gsmic-details}.

\begin{figure}[t]
    \centering
    \includegraphics[width=0.49\textwidth]{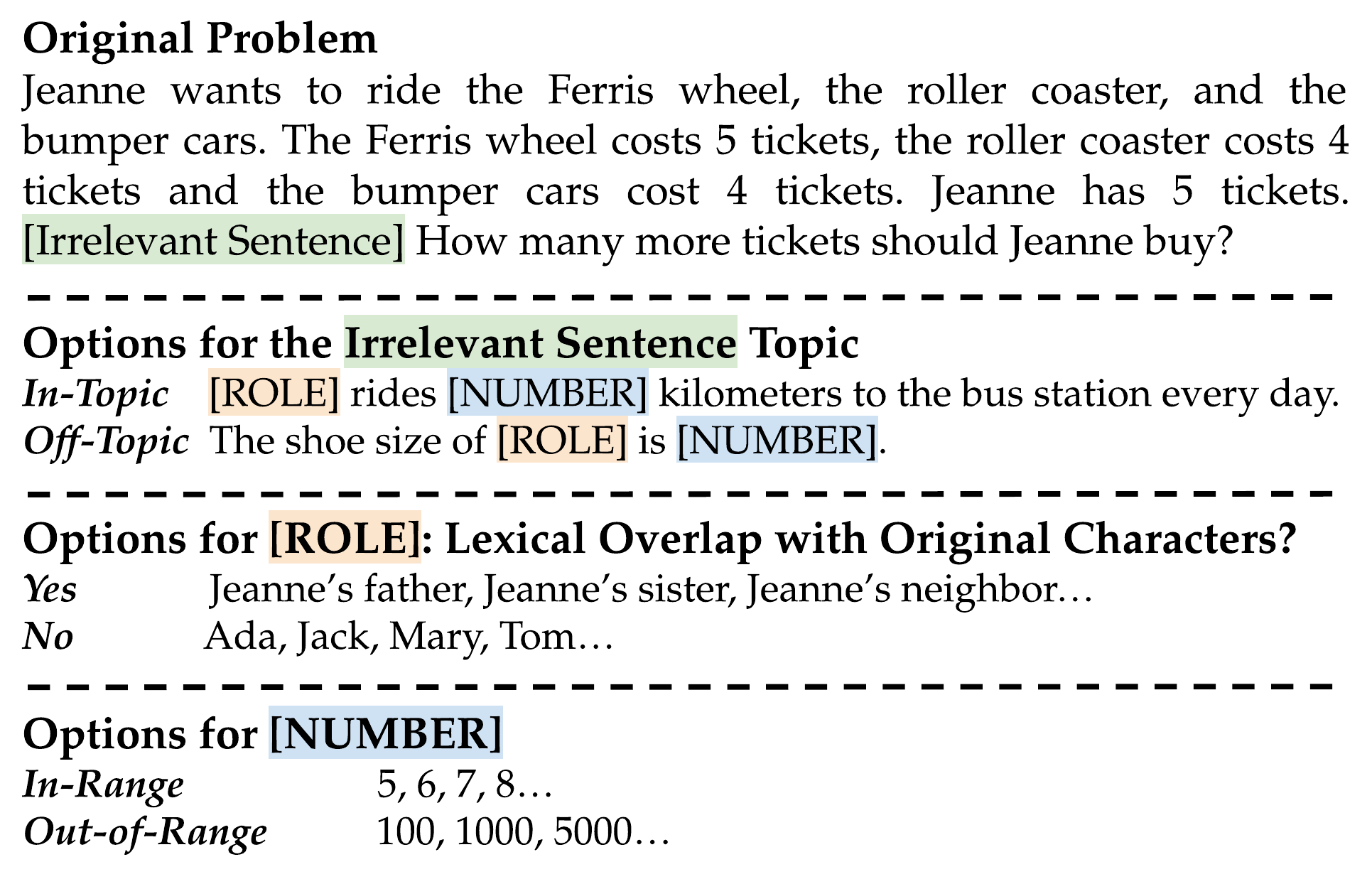}
    \vspace{-15pt}
    \caption{
    Illustration of the considered factors when creating the \noisygsm dataset.
    Best viewed in color.
    }
    \vspace{-10pt}
    \label{fig:controlled-factors}
\end{figure}

We then generate the examples of our new dataset by adding to each base problem one sentence containing irrelevant information.
We use a template-based method (\cref{fig:controlled-factors}) to generate these sentences, which can be characterized by the following three factors:
\begin{itemize}[leftmargin=*,itemsep=0pt,topsep=0pt,parsep=2pt,partopsep=0pt]
    \item \textbf{Topic of the inserted sentence.} We write templates for both in-topic and off-topic sentences.
    In-topic sentences are closely related to the topic of the original problem, whereas off-topic sentences are about a different topic.

    \item \textbf{Role name overlap}. 
    Most sentence templates contain some role name blanks, which can be filled with names that may or may not overlap with the role names that occur in the problem.
    For blank fillers that have overlap with original role names, we: (1) randomly pick a role name \texttt{A} from the original problem description and (2) create the blank fillers with template such as \texttt{A's father} and \texttt{A's sister}.
    
    \item \textbf{Range of numbers}. 
    Since we focus on arithmetic reasoning, most sentence templates also contain a number blank.
    We can choose to fill in the number blank with a number of similar or different magnitude to those in the original problem description.
    Concretely, for a number $a$, if there exists a number $b$ in the original problem description or solution such that $\frac{1}{10} \leq \frac{a}{b} \leq 10$, we consider $a$ as an in-range number, and otherwise an out-of-range number. 
    Since the standard answer to GSM8K problems are all positive integers, we only consider positive integers as the number blank fillers. 
\end{itemize}

We manually verify that (1) all the generated sentences are acceptable in English and that (2) adding them does not affect the standard solution of the base problem.
Because the above factors are orthogonal, we generate for each base example a set of derived examples with different factor combinations. 
The full \noisygsm benchmark consists of 58,052 examples.
More details about the dataset creation process can be found in \cref{sec:appendix-gsmic-details}.

\subsection{Evaluation Metrics}
\label{sec:metrics}
For a problem $p$, we denote its standard solution by $s(p)$, and the solution of method $\mathcal{M}$ by $\mathcal{M}(p)$.
To evaluate the distractibility of $\mathcal{M}$, we consider the following two metrics:
\begin{itemize}[leftmargin=*,itemsep=0pt,topsep=0pt,parsep=2pt,partopsep=0pt]
\item \textbf{Micro accuracy} $\textit{Acc}_\textit{micro}(\mathcal{M}; \mathcal{P})$ is the average accuracy of method $\mathcal{M}$ over all the test problems $\mathcal{P}$.
\begin{align*}
    \textit{Acc}_\textit{micro}(\mathcal{M}; \mathcal{P}) = 
        \frac{
            \sum_{p \in \mathcal{P}} 
                \mathbbm{1}\left[
                    \mathcal{M}(p) = s(p)
                \right]
        }{
            |\mathcal{P}|
        }
\end{align*}
 This means that the micro accuracy weighs all the individual test problems equally.

\item \textbf{Macro accuracy} $\textit{Acc}_\textit{macro}(\mathcal{M}; \mathcal{B})$ is the average accuracy of method $\mathcal{M}$ over classes of test problems, where each class $\mathcal{P}(b)$ consists of the set of test examples derived from the base example $b \in \mathcal{B}$. We define $\mathcal{M}$'s prediction for a class $\mathcal{P}(b)$ to be correct if and only if $\mathcal{M}$'s prediction for all problems in this class are correct.
\begin{align*}
    \textit{Acc}_\textit{macro}(\mathcal{M}; \mathcal{B}) = 
        \frac{
            \sum_{b\in \mathcal{B}}
                \mathbbm{1}\left[
                    \bigwedge_{p\in \mathcal{P}(b)}\left[
                        \mathcal{M}(p) = s(p)
                    \right]
                \right]
        }{
            |\mathcal{B}|
        }
\end{align*}
This means that the macro accuracy is the fraction of base problems that can be consistently solved no matter what irrelevant sentence is being added.

\item \textbf{Normalized accuracy} measures how a method is affected by the distractors, considering its accuracy on base problems.
For a micro or macro accuracy $a_\mathcal{M}$ achieved by method $\mathcal{M}$, we calculate its corresponding normalized accuracy by
\begin{align*}
\textit{norm}(a_\mathcal{M}; \mathcal{M}) = \frac{a_\mathcal{M}}{n_\mathcal{M}},
\end{align*}
where $n_\mathcal{M}$ denotes the base problem accuracy of method $\mathcal{M}$ (\cref{tab:base-accu}).

\end{itemize}

\section{Investigated Solutions}
\begin{figure*}[t!]
    \centering
    \begin{minipage}{0.98\columnwidth}
    \centering
    \includegraphics[width=0.95\textwidth]{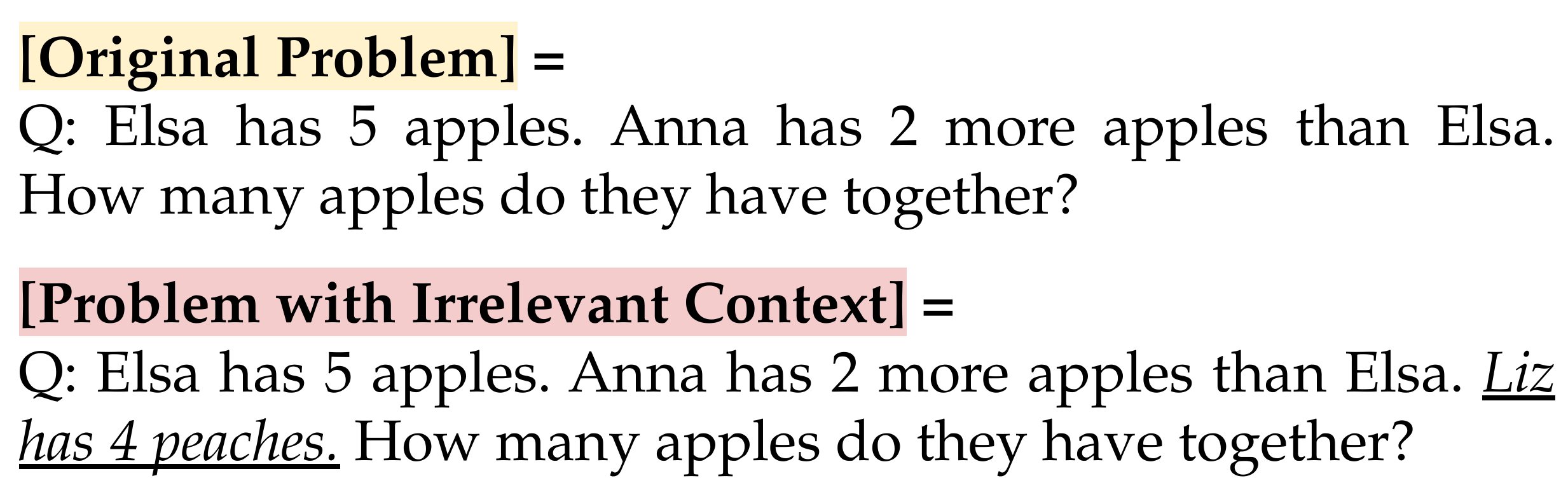}
    \includegraphics[width=0.95\textwidth]{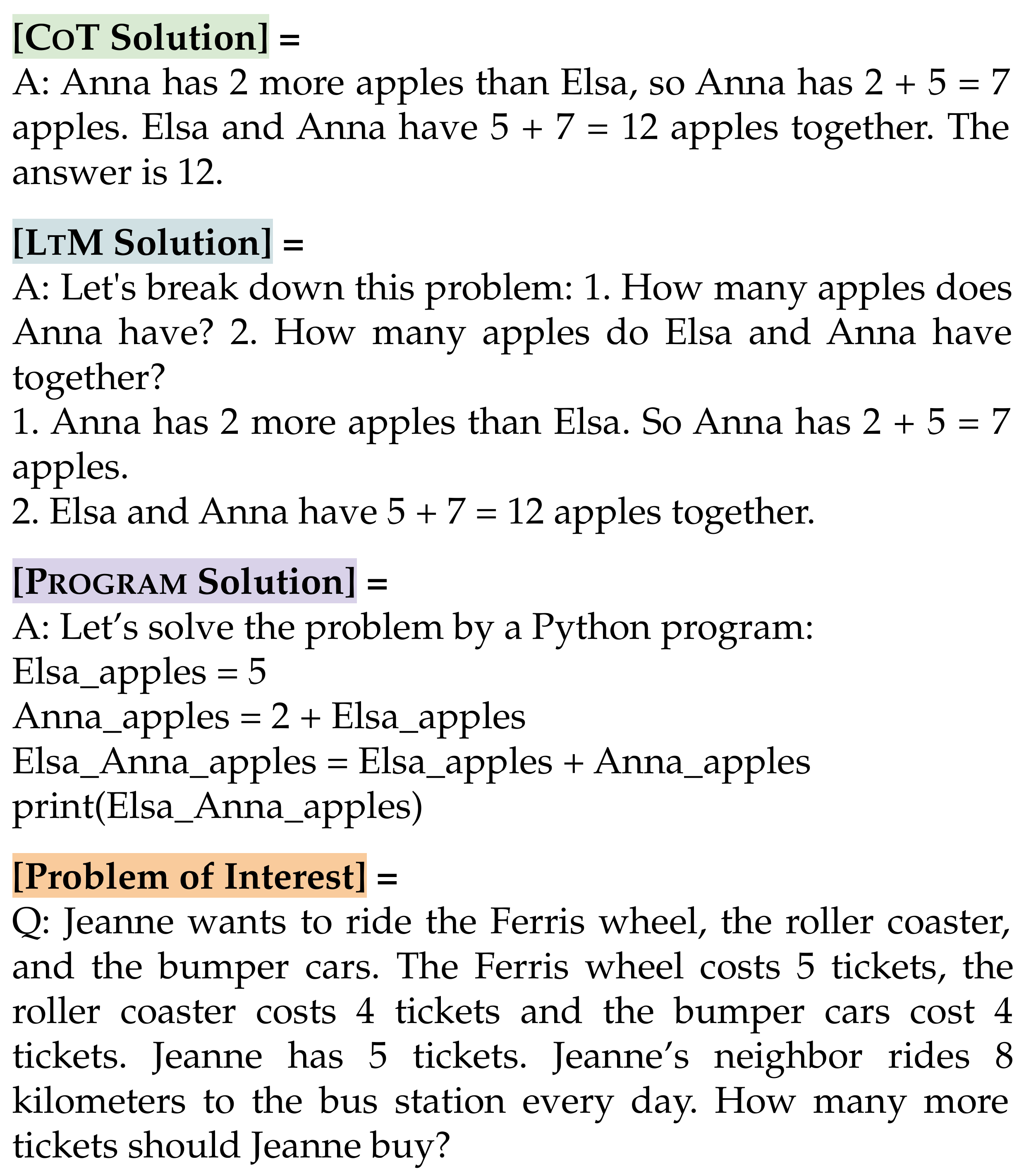}
    \end{minipage}\qquad
    \begin{minipage}{0.98\columnwidth}
    \centering
    \includegraphics[width=\textwidth]{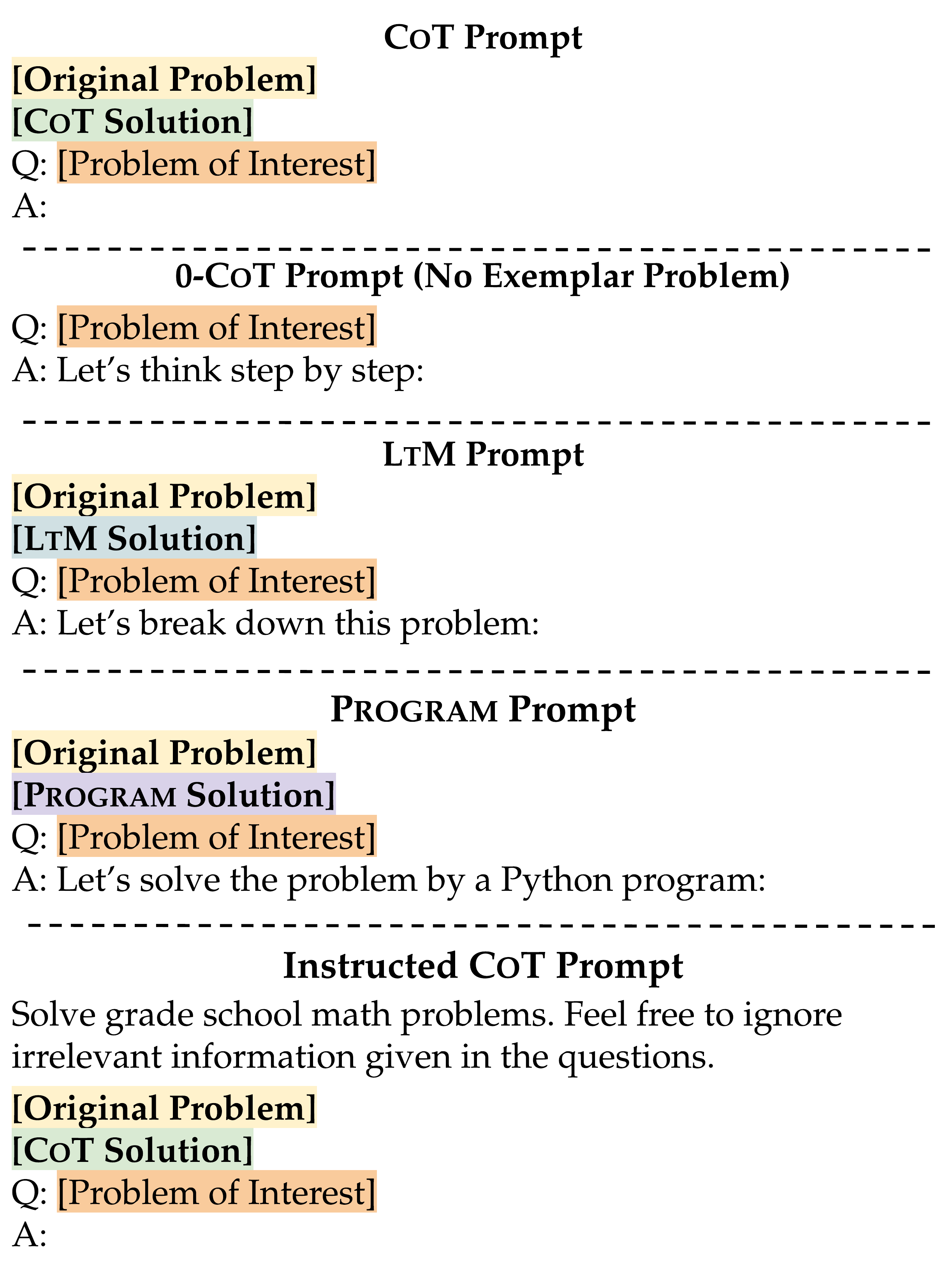}
    \end{minipage}
    \caption{
    Prompt formats for the investigated techniques on the right, which are constructed from building blocks on the left (best viewed in color). 
    The [Problem with Irrelevant Context] is obtained by adding an irrelevant sentence (\textit{\underline{italic and underlined}}) to the original problem description and it can be used as an alternative to the [Original Problem] in the prompts on the right.
    In these prompts, identifiers highlighted and wrapped by brackets (e.g., [Problem of Interest]) are replaced by the contents of the corresponding building blocks. 
    The prompts for all settings can be found in \cref{sec:appendix-prompt}. 
    }
    \label{fig:prompt-design}
\end{figure*}
In the following section, we review the investigated prompting techniques (\cref{subsec:base-techniques}), present the formats of our prompts (\cref{subsec:prompt-design}), and introduce instructed prompting (\cref{subsec:instructions}).
\subsection{Base Techniques}
\label{subsec:base-techniques}
\noindent\textbf{Chain-of-thought prompting \citep[\chainofthought;][]{wei2022chain}} is a prompting technique that guides the language models to solve a problem in a step-by-step manner.
By presenting exemplars that solve the corresponding problems with intermediate reasoning steps in the prompts, \chainofthought significantly improves the reasoning performance over direct answer prediction without such intermediate reasoning steps.

\noindent\textbf{Zero-shot chain-of-thought prompting \citep[0-\chainofthought;][]{kojima2022large}} is a variation of \chainofthought where the prompt does not contain any exemplar. Instead, the model is prompted directly with the problem of interest followed by the instruction ``\textit{Let's think step by step:}''.

\noindent\textbf{Least-to-most prompting \citep[\leasttomost;][]{zhou2022least}} teaches language models to (1) break down a problem into subproblems, and (2) solve those subproblems sequentially using \chainofthought. The final answer is that to the last subproblem.

\noindent\textbf{Program prompts \citep[\program;][]{chowdhery2022palm}} represent the arithmetic reasoning process as a program. Following prior work on solving GSM8K problems with code \citep{chowdhery2022palm,gao2022pal,chen2022program}, we include a Python program as the problem solution in the prompt, and execute the generated Python code using an external Python interpreter to obtain the final answer.

\textbf{Self-consistency} \citep[\selfconsistency;][]{wang2022self,shi2022natural} may further boost the reasoning performance by marginalizing over intermediate reasoning steps that share the same final result.
In practice, SC can be implemented by (1) sampling several solutions from the large language model and (2) taking the majority vote.
Note that SC is orthogonal to above techniques, and can be combined with any of them. 

\subsection{Prompt Design}
\label{subsec:prompt-design}
We present some example prompts used in our experiments (\cref{fig:prompt-design}).
For few-shot prompting techniques (i.e., \chainofthought, \leasttomost and \program), the input prompt includes exemplar problems and their solutions before the problem of interest. 
In order to keep simplicity and avoid over-fitting in prompt engineering, we follow \citet{zhou2022least} on exemplar creation; that is, we only use one simple exemplar for our main experiments. 
This exemplar is either based on the [Original Problem] or the [Problem with Irrelevant Context], which allows us to investigate the effect of irrelevant information in the prompt exemplar.
For 0-\chainofthought, we adhere to \citet{kojima2022large} and directly present the problem of interest followed by ``\textit{A: Let's think step by step:}''.

\subsection{Instructed Prompting}
\label{subsec:instructions}
\iffalse
\begin{figure}[t]
    \centering
    \includegraphics[width=0.48\textwidth]{rawfigs/detailed-prompts-with-instructions.pdf}
    \vspace{-15pt}
    \caption{
    Prompts with instructions (best viewed in color), which requires the language model to ignore irrelevant contexts in natural language.
    Detailed contents for each identifier correspond to those presented in \cref{fig:prompt-design}. \xinyun{Just need to show instruction + CoT, and mention that all prompts use the same instruction. So this figure can be removed and merged into Figure \ref{fig:prompt-design}.}
    }
    \label{fig:instructed-prompts}
\end{figure}
\fi
In addition to presenting irrelevant information in the exemplars, we also investigate whether natural language instructions help language models ignore irrelevant context and become less distracted.
Extending the line of work \citep{suzgun2022challenging,sanh2021multitask,ouyang2022training} that includes a general task description before exemplars, we add the sentence \textit{``Solve grade school math problems. Feel free to ignore irrelevant information given in the questions.''} before our exemplars in the prompt (\cref{fig:prompt-design}), which explicitly \textit{instructs} the language model to ignore irrelevant information in the problem description. 
\section{Experiments}
\begin{table*}[!t]
    \centering \small
    \begin{tabular}{l|cccc|cccc}
    \toprule
    \multirow{2}{*}{\bf Method} & \multicolumn{4}{c|}{\bf Micro Accuracy} & \multicolumn{4}{c}{\bf Macro Accuracy} \\
    & 2 Steps & $>$2 Steps & Overall & \it Norm & 2 Steps  & $>$2 Steps & Overall & \it Norm\\
    \midrule 
    \multicolumn{9}{l}{\textit{Prompting Exemplar w/o Irrelevant Context,} \texttt{code-davinci-002}} \\
    \cdashlinelr{1-9}
    \chainofthought & 73.5 & 70.8 & 72.4 & \it 76.2 & 8.3 & 2.5 & 6.0 & \it 6.3 \\
    \chainofthought + \textsc{Inst.} &  79.0 & 76.0 & 77.8 & \it 81.8 & 20.0 & 7.0 & 15.0 & \it 15.8 \\   
    0-\chainofthought & 29.0  & 29.1 & 29.0 & \it 65.9 & 1.7 & 0.0 & 1.0 & \it 2.3 \\
    0-\chainofthought+\textsc{Inst.} & 31.6 & 28.8 & 30.5 & \it 69.3 & 1.7 & 0.0 & 1.0 & \it 2.3 \\
    \leasttomost & 74.9  & \bf 81.5  & 77.5 & \it 82.4 & 16.7 & 20.0 & 18.0 & \it 19.1 \\
    \leasttomost + \textsc{Inst.} & \bf 80.1 & 81.3 & \bf 80.6 & \textbf{\textit{85.7}}& \bf 18.3 & \bf 35.0 & \bf 25.0 & \textbf{\textit{26.6}} \\
    \program & 59.1 & 47.4 & 54.4 & \it 65.5 & 6.7 & 2.5 & 5.0 & \it 6.0 \\
    \program + \textsc{Inst.} & 60.6 & 50.9 & 56.7 & \it 68.3 & 6.7 & 5.0 & 6.0 & \it 7.2 \\
    \midrule 
    \chainofthought+ SC & 87.6  & 90.1 & 88.1 & \it 91.8 & 29.0 & 28.3 & 30.0 & \it 31.3 \\
    0-\chainofthought + SC & 61.6 & 68.4 & 64.3 & \it 84.6 & 0.0 & 2.5 & 1.0 & \it 1.3 \\
    \leasttomost+ SC & \bf 92.4 & \bf 94.8 & \bf 93.4 & \textbf{\textit{94.3}} & \bf 51.6 & \bf 35.0 & \bf 45.0 & \bf \textbf{\textit{45.5}} \\
    \program+ SC & 73.5 & 76.1 & 74.6 & \it 82.0 & 16.7 & 7.5 & 13.0 & \it 14.3 \\
    \midrule 
    \multicolumn{9}{l}{\textit{Prompting Exemplar w/o Irrelevant Context,} \texttt{text-davinci-003}} \\
    \cdashlinelr{1-9}
    \chainofthought & 69.3 & 66.9 & 68.4 & \it 85.4 & 10.0 & 7.5 & 9.0 & \it 11.3 \\
    \chainofthought + \textsc{Inst.} & 72.0 & 70.3 & 71.3 & \it 89.1 & \bf 11.7 & \bf 12.5 & \bf 12.0 &\textit{\textbf{15.0}} \\
    \leasttomost & 78.0 & \bf 73.6 & 76.3 & \it 94.2 & 5.0 & 0.0 & 5.0 & \it 6.2 \\
    \leasttomost + \textsc{Inst.} & \bf 80.5 & 70.9 & \bf 76.7 & \textbf{\textit{94.7}} & 5.0 & 0.0 & 5.0 & \it 6.2 \\
    \midrule
    \multicolumn{9}{l}{\textit{Prompting Exemplar w/ Irrelevant Context,} \texttt{code-davinci-002}} \\
    \cdashlinelr{1-9}
    \chainofthought &  79.8  & 72.4 & 76.8 & \it 80.8 & 16.7 & 10.0 & 14.0 & \it 14.7 \\
    \chainofthought + \textsc{Inst.} & 80.5 & 74.4 & 78.1 & \it 82.2 & 20.0 & 12.0 & 17.0 & \it 17.9 \\
    \leasttomost    & 78.1  & 84.6 & 80.7 & \it 85.9 & \bf 23.3 & \bf 35.0 & \bf 28.0 & \textbf{\textit{29.8}}\\
    \leasttomost + \textsc{Inst.} & \bf 81.0 & \bf 85.4 & \bf 82.8 & \textbf{\textit{88.1}} & \bf 23.3 & \bf 35.0 & \bf 28.0 & \textbf{\textit{29.8}} \\
    \program     & 67.0 & 55.0 & 62.2 & \it 74.9 & 11.7 & 5.0 & 9.0 & \it 10.8 \\
    \program + \textsc{Inst.} & 68.8 & 54.8 & 63.2 & \it 76.1 & 15.0 & 7.5 & 12.0 & \it 14.5 \\
    \bottomrule
    \end{tabular}
    \caption{
    Micro and macro accuracies ($\times 100$) on the \noisygsm-4K dataset. 
    SC denotes self-consistency. \textit{Norm} is the overall accuracy normalized by the fraction of solved base problems (\cref{tab:base-accu}), which is a measure for robustness w.r.t. irrelevant information. For \texttt{text-davinci-003}, the base problem accuracy with \chainofthought is 80.0, and the base problem accuracy with \leasttomost is 81.0.
    The best numbers in each column for each section (i.e., whether using \texttt{code-davinci-002} or \texttt{text-davinci-003}, whether using exemplar with irrelevant context or not, and whether using self-consistency or not) are in \textbf{boldface}. 
    }
    \label{tab:main-results}
\end{table*}

Being mindful of the experiment costs, we uniformly sample 4,000 examples from the \noisygsm dataset (denoted by \noisygsm-4K)\footnote{Our sampled \noisygsm-4K covers all 100 base problems.} for evaluation and analysis purposes throughout this paper.
Unless otherwise specified, we mainly use \texttt{code-davinci-002} in our experiments, and we also evaluate \texttt{text-davinci-003} which is a model trained with RLHF to better follow instructions \citep{ouyang2022training}.
For experiments without self-consistency decoding, we use greedy decoding (i.e., temperature $\tau=0$); for self-consistency experiments that require multiple samples for a problem, we sample 20 responses with temperature $\tau=0.7$ following \citet{wang2022self}.

\subsection{Main Results on \noisygsm}
\label{sec:main-results}
\begin{figure}[t]
    \centering
    \includegraphics[width=0.46\textwidth]{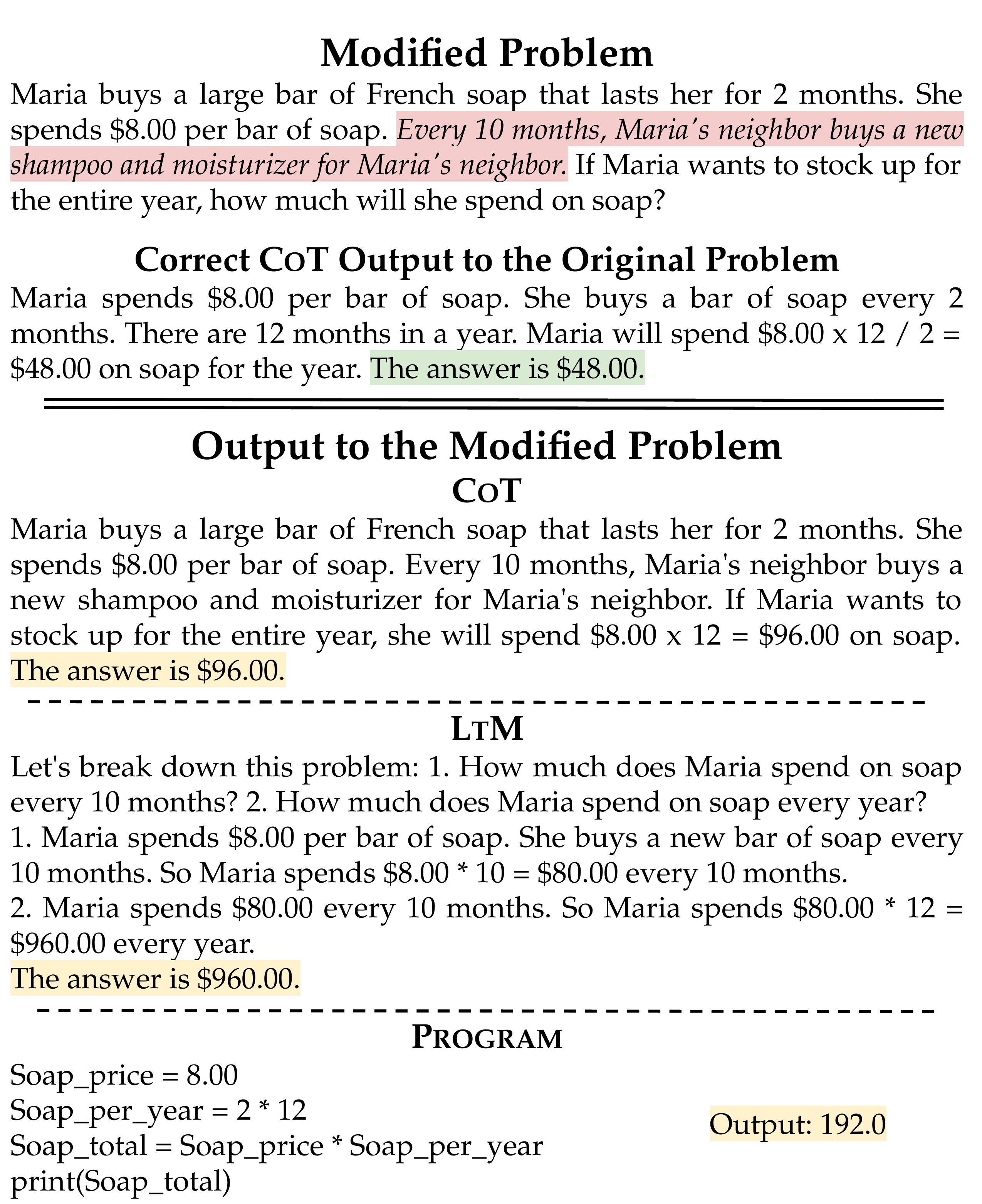}
    \caption{
    Example problem and corresponding outputs by different prompting techniques (best viewed in color). 
    The \chainofthought answer to the original problem is highlighted in green. 
    The added irrelevant sentence is in \textit{italic and highlighted in red}, which causes different errors (highlighted in yellow) for all prompting techniques. 
    More examples of model predictions can be found in \cref{sec:appendix-sample-output}.
    }
    \vspace{-8pt}
    \label{fig:example-main}
\end{figure}

We compare the performance of different prompting techniques on \noisygsm-4K  (\cref{tab:main-results}), in terms of both micro and macro accuracies, as well as their corresponding normalized accuracies.
Overall, we observe significant performance drop for both models with all prompting techniques. 
The drop on macro accuracy is especially large, showing that fewer than 30\% of the base problems are consistently solved after adding distractors. Comparing the results of two models, \texttt{text-davinci-003} achieves better normalized micro accuracy than \texttt{code-davinci-002}, though its macro accuracy is mostly worse. In \cref{fig:example-main}, we present a \noisygsm-4K example where a single irrelevant sentence causes different types of errors in investigated prompting techniques. One common error type is wrongly using the number in the irrelevant sentence, as shown in the \leasttomost prediction and other examples in \cref{sec:appendix-sample-output}.  Even if the model does not directly use the irrelevant number for numerical calculation, the presence of the irrelevant sentence in the reasoning steps alone can still cause a wrong prediction, as shown in the \chainofthought prediction.

\noindent\textbf{\leasttomost is generally the most robust technique to irrelevant context.}
In terms of micro accuracy, \leasttomost outperforms all other prompting methods across models. 
Using \texttt{code-davinci-002}, \leasttomost achieves about double macro accuracy of \chainofthought. Interestingly, with \texttt{text-davinci-003}, despite that \leasttomost outperforms \chainofthought on the micro accuracy, its macro accuracy is lower. Specifically, \texttt{text-davinci-003} is highly susceptible to irrelevant context with role overlap; e.g., such irrelevant sentences decrease the macro accuracy to 0 on problems with more than 2 reasoning steps. See Table~\ref{tab:breakdown} for the breakdown performance on different types of irrelevant context.

\noindent\textbf{Selecting exemplars with distractors mitigates the distractibility.} 
For few-shot prompts, we find that using exemplars with distractors (i.e., including problems with irrelevant context) consistently outperforms using the original exemplars without distractors across prompting techniques. While prior work has shown that training and fine-tuning with different types of problems improves model robustness \citep{li2022large}, our results show that prompting with exemplars that demonstrate how to ignore irrelevant context also results in significant robustness improvement. In \cref{tab:instructed-on-original}, we further show that using exemplars with distractors does not cause a performance drop on the original GSM8K dataset, indicating that such a prompt design can be beneficial in achieving better accuracy and robustness simultaneously.
%While our results are in line with \citet{min2022rethinking} that out-of-domain exemplars lead to decent performance, we also advocate for the approach of using in-domain exemplars whenever possible in order to achieve higher performance.

\noindent\textbf{Self-consistency significantly reduces the distractibility.} 
Taking the majority vote from 20 samples,\footnote{If there is a tie, we take a random top-tier result for evaluation, following \citet{wang2022self} and \citet{shi2022natural}.} SC improves the overall micro accuracy by more than 11  percentage points.
This means that in addition to improving model performance on clean arithmetic reasoning tasks \citep{wang2022self}, SC also substantially reduces the distractibility of large language models to irrelevant context.
The gain on micro accuracy is notably large on 0-\chainofthought (35.5 percentage points).
Furthermore, the correct answer for 99.7\% of the problems is in the 20 sampled answers for both \chainofthought and \leasttomost. 
Even for 0-\chainofthought, the recall of correct solutions within 20 samples is 96.5\%. 
Despite these improvements, the best macro accuracy among all prompting techniques is only $45\%$, suggesting that for more than half of the base problems, SC fails to prevent the model from being distracted by different variants of irrelevant information.
These results imply that a better algorithm may be developed to further reduce the distractibility based on a few sampled solutions.

\subsection{Break-Down Analysis}
\begin{table*}[t!]
    \centering \small
    \begin{tabular}{c|cccccc|cccccc}
         \toprule
         \multirow{3}{*}{\bf Method} & \multicolumn{6}{c|}{\bf Micro Accuracy} & \multicolumn{6}{c}{\bf Macro Accuracy} \\
         \cmidrule(lr){2-7} \cmidrule(lr){8-13}
         & \multicolumn{2}{c}{\bf Topic} & \multicolumn{2}{c}{\bf Role Overlap} & \multicolumn{2}{c|}{\bf Num. Range} & \multicolumn{2}{c}{\bf Topic} & \multicolumn{2}{c}{\bf Role Overlap} & \multicolumn{2}{c}{\bf Num. Range}\\
         \cmidrule(lr){2-3} \cmidrule(lr){4-5} \cmidrule(lr){6-7}
         \cmidrule(lr){8-9} \cmidrule(lr){10-11} \cmidrule(lr){12-13}
          & In & Off & Yes & No & In & Out & In & Off & Yes & No & In & Out \\
         \midrule
         \multicolumn{13}{l}{\textit{Prompting Exemplar w/o Irrelevant Context (\texttt{code-davinci-002})}} \\
         \cdashlinelr{1-13}
         \chainofthought & 63.1 & 80.7 & 68.3 & 76.6 & 70.2 & 74.6 & 10.2 & 33.0 & 10.3 & 22.2 & 11.0 & 19.0 \\
         \leasttomost & \bf 70.8 & \bf 83.4 & \bf 77.0 & \bf 78.2 & \bf 77.2 & \bf 77.8 & \bf 23.5 & \bf 45.0 & \bf 25.8 & \bf 35.4 & \bf 27.0 & \bf 29.0 \\
         \program & 44.1 & 63.5 & 50.7 & 58.4 & 54.3 & 54.5 & 4.1 & 24.0 & 9.3 & 16.2 & 7.0 & 11.0 \\
         \midrule
         \multicolumn{13}{l}{\textit{Prompting Exemplar w/o Irrelevant Context (\texttt{text-davinci-003})}} \\
         \cdashlinelr{1-13}
         \chainofthought & 63.3 & 72.9 & 68.7 & 68.1 & 67.2 & 69.6 & \bf 16.3 & \bf 36.0 & \bf 17.5 & \bf 20.2 & \bf 19.0 & \bf 22.0 \\
         \leasttomost & \bf 75.4 & \bf 76.9 & \bf 75.6 & \bf 76.8 & \bf 75.3 & \bf 77.2 & 6.1 & 7.0 & 6.2 & 9.1 & 6.0 & 6.0 \\
         \midrule 
         \multicolumn{13}{l}{\textit{Prompting Exemplar w/ Irrelevant Context (\texttt{code-davinci-002})}} \\
         \cdashlinelr{1-13}
         \chainofthought & 70.2 & 82.7 & 73.6 & 80.2 & 76.1 & 77.7 & 18.4 & 43.0 & 21.6 & 32.3 & 22.0 & 26.0 \\
         \leasttomost & \bf 73.0 & \bf 87.5 & \bf 81.4 & \bf 80.2 & \bf 80.0 & \bf 81.4 & \bf 28.6 & \bf 58.0 & \bf 37.1 & \bf 42.4 & \bf 41.0 & \bf 35.0\\
         \program & 52.9 & 70.5 & 60.2 & 64.5 & 61.5 & 62.8 & 10.2 & 37.0 & 14.4 & 23.2 & 15.0 & 17.0\\
         \bottomrule
    \end{tabular}
\caption{
    Breakdown accuracies ($\times 100$) w.r.t. the factors of the added irrelevant sentence.
    Lower accuracy indicates the model is more fragile to the corresponding type of irrelevant contexts.
    Note that the macro average accuracies are higher than the corresponding ones reported in \cref{tab:main-results}, as we only include a subset of created problems (i.e., those corresponding to the appropriate factor) here to compute the metric.
    The best result in each column is in \textbf{boldface}.}
\label{tab:breakdown}
\end{table*}

\subsubsection{Factors of the Irrelevant Context}
We analyze the performance of \chainofthought, \leasttomost and \program with respect to the considered factors (\cref{sec:dataset}) of the irrelevant sentences (\cref{tab:breakdown}). For both models, we find that (1) in-topic sentences with (2) role name overlap and (3) in-range numbers are generally more challenging, which is examplified by \cref{fig:example-main}.
For \leasttomost, the latter two factors do not have a large effect on the micro accuracy. The difference is more significant for the macro accuracy and, as an anomaly, using distractors with in-range numbers turns out to be less challenging than out-of-range numbers when using irrelevant context in the exemplar.
Again, with \texttt{code-davinci-002}, \leasttomost outperforms \chainofthought and \program on all investigated sub-categories. On the other hand, using \texttt{text-davinci-003}, \leasttomost outperforms \chainofthought in terms of the micro accuracy, but the macro accuracy is much lower on all sub-categories.

\subsubsection{Break-Down Accuracies w.r.t. \# Steps}
\begin{figure}[t]
    \centering
    \includegraphics[width=0.5\textwidth]{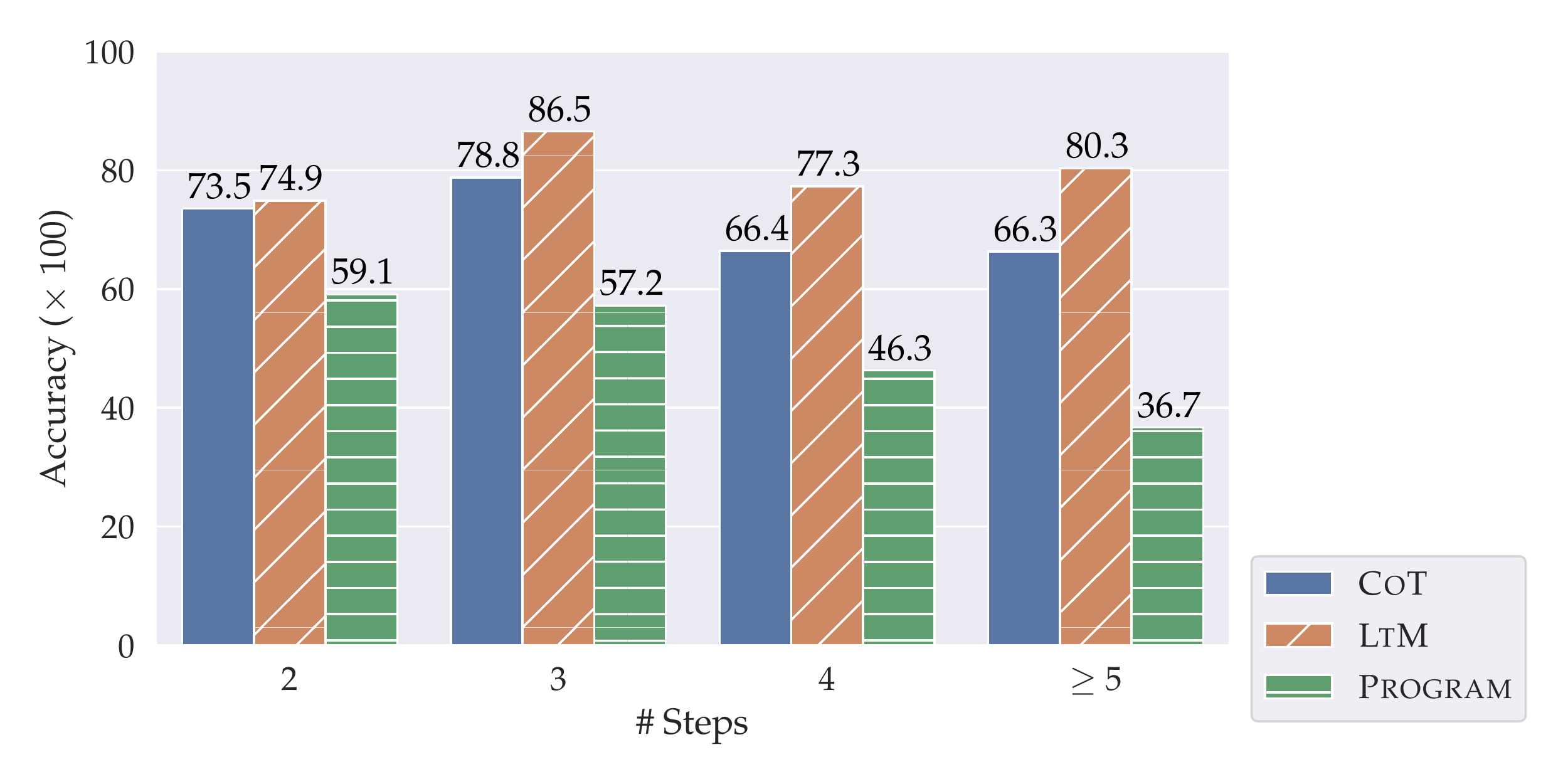}
    \vspace{-15pt}
    \caption{Micro accuracies on \noisygsm-4K with respect to the number of required reasoning steps. }
    \label{fig:breakdown-accu}
    \vspace{-1em}
\end{figure}

We analyze the break-down accuracies for problems with respect to the reasoning steps (\cref{fig:breakdown-accu}).
While we see a significant drop for \chainofthought and \program on problems that require four or more steps in the reasoning process, the performance of \leasttomost is fairly consistent across difficulty. 
In addition to the advantage of \leasttomost on clean problems for complicated reasoning \citep{zhou2022least}, our results show that \leasttomost is also less sensitive to irrelevant context for complicated problems that require more steps to solve.

\subsection{Instructed Prompting Improves Robustness to Irrelevant Context}
We have shown that using exemplars with distractors improves robustness to irrelevant context. 
We also compare the performance of instructed prompting and that of the prompts without instructions in \cref{tab:main-results}.
Adding instructions to \chainofthought, \leasttomost, and \program consistently improves their performance.
% We find that the prompts with instructions consistently reach the best performance for \chainofthought, \leasttomost, and \program. 
Surprisingly, instructed prompting with original exemplars reaches comparable or even better performance than uninstructed prompting that uses exemplars with distractors for both \chainofthought and \leasttomost.
Note that adding the instruction \textit{``Solve grade school math problems.''} alone does not significantly improve the performance, and it is the instruction \textit{``Feel free to ignore irrelevant information given in the questions.''} that makes the difference.
Similar to the instruction \textit{``Let's think step by step.''} employed by 0-CoT, this shows that language models are---to some extent---able to follow natural language instructions in a way that dramatically changes their problem solving behavior, suggesting that such instructions may be useful for guiding the behavior of language models on more tasks.

\begin{table}[t]
    \centering \small
    \begin{tabular}{cccc}
    \toprule
    \multirow{2}{*}{\bf Method}  & \bf Exemplar & \multicolumn{2}{c}{\bf Accuracy} \\
    & \textsc{w/ IrrCtx?} &  GSM8K Dev.  & SVAMP Test  \\
    \midrule
    \chainofthought & \cmark & 59.3 & 79.1 \\
        & \xmark             & \underline{60.3} & \underline{77.6}  \\
    \cdashlinelr{1-4}
    \chainofthought & \cmark & 59.3 & 79.1\\
    \quad + \textsc{Inst.}& \xmark & 58.8 & 78.7\\
    \midrule
    \leasttomost    & \cmark   &  61.9    &  76.9\\
                    & \xmark   &  \underline{59.8}  &  \underline{76.6}\\
    \cdashlinelr{1-4}
    \leasttomost & \cmark &  60.9        & 76.2\\
    \quad + \textsc{Inst.} & \xmark  & 60.3  & 76.3\\
    \midrule
    \program    & \cmark   &  58.6    & 80.0  \\
                    & \xmark   &  \underline{59.8}  & \underline{77.3} \\
    \cdashlinelr{1-4}
    \program & \cmark &  59.2        & 77.9 \\
    \quad + \textsc{Inst.} & \xmark  & 61.1  & 77.8 \\
    \bottomrule
    \end{tabular}
    \caption{
    Accuracies ($\times 100$) on the GSM8K development set and the SVAMP test set.
    \textsc{IrrCtx} denotes irrelevant contexts, and +\textsc{Inst.} denotes instructed prompting.
    The baseline results (i.e., those using the simplest exemplars without irrelevant context and without instructions) are \underline{underlined}.
    }
    \label{tab:instructed-on-original}
\end{table}
On the original GSM8K development set \citep{cobbe2021training,zhou2022least}, we do not observe a drop in accuracy when using exemplars with irrelevant information, adding natural language instructions, or both (\cref{tab:instructed-on-original}).
The same holds for SVAMP \citep{patel2021nlp}, an arithmetic reasoning benchmark constructed by applying different types of variations to math problems from existing clean datasets, e.g., changing sentence structures, asking different questions with the same information, etc.
This is impressive because the results on \noisygsm show that prompt exemplars with irrelevant information and instructed prompting both improve robustness. For the \program prompt, we find that using exemplars with distractors even increases performance on SVAMP.

\vspace{-0.5em}
\subsection{Complicated Prompts May Hurt the Robustness to Irrelevant Context}
\begin{table}[!t]
    \centering \small
    \begin{tabular}{ccccc}
    \toprule
    \multirow{2}{*}{\bf Method}  & \bf \#Prompting & \bf GSM8K & \multicolumn{2}{c}{\bf \noisygsm-4K}\\
    &\bf Exemplars &  Dev.  & 2 Steps & $>$ 2 Steps\\
    \midrule
    \chainofthought & 1 & 60.3 & 73.6 & 70.8  \\
        &  4          & 66.3 & 78.0 & 69.4 \\
    \cdashlinelr{1-5}
    \chainofthought & 1 & 58.8 & 79.0 & \bf 76.0 \\
    + \textsc{Inst.} & 4 & \bf 66.5 & \bf 79.2 & 70.6 \\
    \bottomrule
    \end{tabular}
    \caption{
    Micro accuracies ($\times 100$) on the GSM8K development set and \noisygsm-4K.
    \# Prompting exemplars denotes the number of exemplars used in the prompt. 
    The best number in each column is in \textbf{boldface}.
    }
    \label{tab:complicated-less-robust}
\end{table}

We compare our 1-exemplar \chainofthought prompt (\cref{fig:prompt-design}) to a 4-exemplar prompt \citep[Appendix D of][]{zhou2022least}, which is reported as the best-performing \chainofthought prompt on GSM8K, on \noisygsm (\cref{tab:complicated-less-robust}).
Note that the 1-exemplar \chainofthought prompt only includes a problem with a 2-step solution, while the 4-exemplar prompt includes problems that require more reasoning steps.
While the 4-exemplar prompt leads to better performance on the original GSM8K development set, the 4-exemplar prompt is surprisingly more susceptible to the distraction provided by the irrelevant context.
In particular, the 4-exemplar prompt is consistently worse than the 1-exemplar prompt on problems with more than 2 intermediate steps.
Even for 2-step prompts, the accuracy improvement from adding more exemplars is almost negligible when using instructions (79.0 vs 79.2).
Overall, this finding indicates that adding more exemplars can make the prompt less robust as it leads to some overfitting.
% \footnote{The accuracy of the 4-exemplar \chainofthought prompt on the GSM8K development set is higher than reported in \citet{zhou2022least}, because we corrected an error in their prompt. More details are in the supplementary materials.}

\subsection{Extension to DROP}
In addition to GSM-IC, we extend our evaluation to the DROP dataset \citep{dua2019drop}, where the task is to answer a question according to a long passage that naturally contains irrelevant context. We show an example about football games in \cref{tab:drop-ex}.

\begin{table}[t]
    \small \centering
    \scalebox{0.9}{
    \begin{tabular}{lcc}
    \toprule 
    \bf Method & \texttt{code-davinci-002} & \texttt{text-davinci-003} \\
    \midrule 
    \chainofthought & 67.4 & 68.2 \\
    \chainofthought + \textsc{Inst.} & 68.9 & 69.9 \\
    \leasttomost & 73.4 & 70.2 \\
    \leasttomost + \textsc{Inst.} & \bf 74.4 & \bf 72.8 \\
    \bottomrule
    \end{tabular}}
    \caption{Accuracies ($\times 100$) on the football split of DROP \citep{dua2019drop} benchmark.}
    \label{tab:drop-results}
    \vspace{-1em}
\end{table}

\begin{table}[h!]
    \centering
    \begin{tabular}{p{0.45\textwidth}}
    \toprule
    \normalsize \textbf{Paragraph} \\
    The Seahawks played the San Francisco 49ers. In the first quarter, the Hawks RB Julius Jones got a 27-yard TD run, along with DT Craig Terrill returning a fumble 9 yards for a touchdown. In the third quarter, the 49ers almost rallied as RB H. J. Torres made a 12-yard TD pass to Lucas Nelly, along with Mare kicking a 32-yard field goal. In the final quarter, Julius Jones got another 11-yard TD. \\
    \textbf{Question} \\
    How many yards do the shortest touchdown run and the longest touchdown pass combine for? \\
    \normalsize \textbf{\leasttomost solution}\\
    To answer the question "How many yards do the shortest touchdown run and the longest touchdown pass combine for?", we need to know: "How many yards was the shortest touchdown run?", "How many yards was the longest touchdown pass?". \\
    Q: How many yards was the shortest touchdown run? \\
    A: All the touchdown runs are: a 27-yard touchdown run, a 9-yard touchdown run, a 11-yard touchdown run. The smallest number between 27, 9, 11 is 9. The shortest touchdown run was 9 yards. The answer is 9. \\
    Q: How many yards was the longest touchdown pass? \\
    A: All the touchdown passes are: a 12-yard touchdown pass. The longest touchdown pass was 12 yards. The answer is 12. \\
    Q: How many yards do the shortest touchdown run and the longest touchdown pass combine for? \\
    A: We know that the shortest touchdown run was 9-yard. We also know that the longest touchdown pass was 12-yard. 9 + 12 = 21. The answer is 21. \\
    \bottomrule
    \end{tabular}
    \caption{A DROP example about football games.}
    \label{tab:drop-ex}
    \vspace{-2em}
\end{table}

We use the \chainofthought and \leasttomost prompts in \citep{zhou2022least} as the baselines, and we evaluate the prompt variants with the instruction \textit{``Solve following questions. Feel free to ignore irrelevant information given in the questions.''} added before the exemplars. 
Note that by adding a problem reduction step in the exemplar solution, the least-to-most prompt implicitly leads the model to come up with relevant subproblems to solve the given problem. 
Again, we observe that the instruction consistently improves the performance of both \chainofthought and \leasttomost prompting (\cref{tab:drop-results}).

\section{Conclusion and Discussion}
In this work, we introduce \noisygsm, a dataset that supports comprehensive study of the distractibility of large language models when performing arithmetic reasoning in presence of irrelevant contexts. 
We examine a variety of prompting techniques on \noisygsm, and demonstrate that they are all sensitive to the irrelevant information in the problems. 
Among the studied techniques, self-consistency \citep{wang2022self} leads to a substantial improvement in robustness to irrelevant context across the board, and presenting example problems with irrelevant context in the prompt also consistently improves the performance. Similarly, we find that simply adding an instruction to ignore irrelevant information brings notable performance gains on our benchmark.

Despite the improvement achieved by these methods, the fundamental issue remains: a single piece of irrelevant information can distract the models and substantially degrade their performance, even on problems whose clean versions they correctly solve. 
We encourage researchers to also prioritize improving on this fundamental limitation when developing new training and prompting techniques.
We leave further investigation on the distractibility for other tasks and different language models for future work. 

%\km{Alternate framing of the fundamental issue: A single irrelevant piece of information can substantially degrade the perceived ability of LLMs to perform arithmetic reasoning, even on problems whose clean versions they perfectly solve. As newer prompting techniques are developed, we encourage researchers to also prioritize improvements regarding this fundamental limitation.}
% \km{Alternate framing of the fundamental issue: A single irrelevant piece of information can substantially degrade the perceived ability of LLMs to perform arithmetic reasoning, even on problems whose clean versions they perfectly solve. As newer prompting techniques are developed, we encourage researchers to also prioritize improving on this fundamental limitation.}

%In light of recent work on whether large language models learn meanings \citep{bender2020climbing,li2021implicit,shi2022natural}, our results on instructed prompting can be viewed as distant evidence that meanings are encoded in large language models.
% \freda{Follow communication intents.}
% \freda{potential explanation why CoT doesn't work for DROP.}

\section*{Acknowledgement}
We would like to thank Dale Schuurmans, Olivier Bousquet and Jack Nikodem for helpful discussion and feedback.

% In the unusual situation where you want a paper to appear in the
% references without citing it in the main text, use \nocite

\bibliography{main}
\bibliographystyle{icml2023}

%%%%%%%%%%%%%%%%%%%%%%%%%%%%%%%%%%%%%%%%%%%%%%%%%%%%%%%%%%%%%%%%%%%%%%%%%%%%%%%
%%%%%%%%%%%%%%%%%%%%%%%%%%%%%%%%%%%%%%%%%%%%%%%%%%%%%%%%%%%%%%%%%%%%%%%%%%%%%%%
% APPENDIX
%%%%%%%%%%%%%%%%%%%%%%%%%%%%%%%%%%%%%%%%%%%%%%%%%%%%%%%%%%%%%%%%%%%%%%%%%%%%%%%
%%%%%%%%%%%%%%%%%%%%%%%%%%%%%%%%%%%%%%%%%%%%%%%%%%%%%%%%%%%%%%%%%%%%%%%%%%%%%%%
\newpage
\appendix
\onecolumn
\section{\noisygsm Details}
\label{sec:appendix-gsmic-details}
\begin{figure}[ht]
    \centering
    \includegraphics[width=0.48\textwidth]{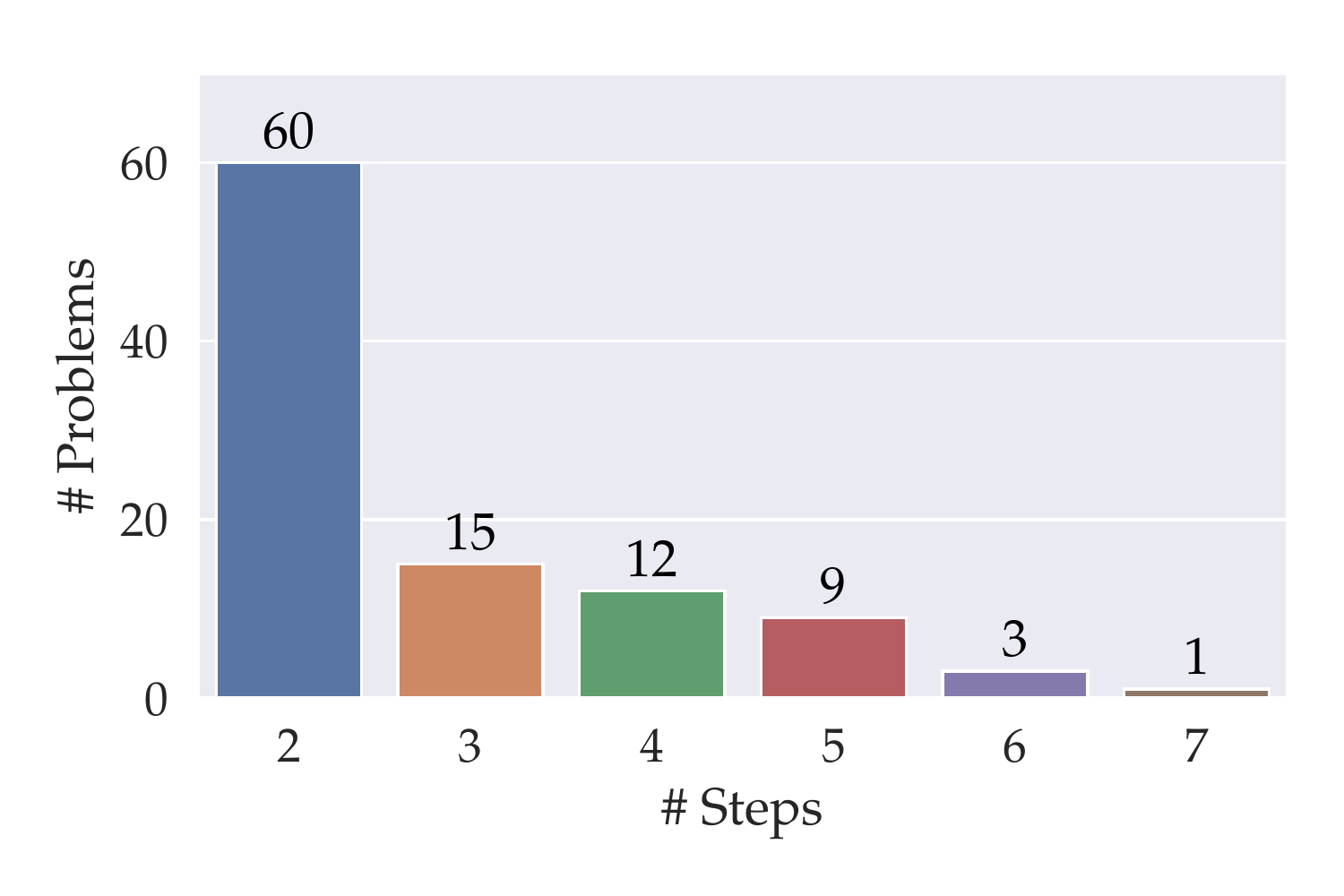}
    \vspace{-15pt}
    \caption{Base problem distribution of \noisygsm with respect to the number of reasoning steps in the ground truth problem solution. }
    \label{fig:n-steps-distribution}
\end{figure}

Each of the 100 base problem require two to seven steps to solve (\cref{fig:n-steps-distribution}). 

Starting from the base problems, we follow the protocols below to create \noisygsm (\cref{sec:dataset}).

\begin{enumerate}
    \item \textbf{Irrelevant sentence template}. 
    \begin{enumerate}
        \item For in-topic sentences, we manually write templates within the topic that is close to the original problem description. 
        We are particularly careful about the shareable stuff, for example, money is sometimes considered shareable between family members. 
        In such cases, we make sure that the added do not change the amount of shareable stuff to ensure that the final standard answer is not affected. 
        \item For off-topic sentences, we use general templates (\cref{tab:appendix:off-topic-templates}) for all problems unless some of them can be considered as in-topic sentences for some problems---for example, the sentence ``\textit{The height of \{role\} is \{number\} feet.}'' is considered as an in-topic sentence for problems about heights of people.
        \begin{table}[h]
            \centering
            \begin{tabular}{l}
                 \toprule
                 The shoe size of \texttt{[ROLE]} is \texttt{[NUMBER]}.\\
                 \texttt{[ROLE]} is \texttt{[NUMBER]} years old. \\
                 The height of \texttt{[ROLE]} is \texttt{[NUMBER]} feet. \\
                 \texttt{[ROLE]} bought \texttt{[NUMBER]} tomatoes from the grocery store. \\
                 \texttt{[ROLE]} has read \texttt{[NUMBER]} books in the past year. \\
                 \bottomrule
            \end{tabular}
            \caption{Off-topic sentence templates for \noisygsm.}
            \label{tab:appendix:off-topic-templates}
        \end{table}
        \item We make sure that all sentences derived by each template are grammatical English sentences. 
        \item We write four in-topic and choose four off-topic distractor sentence templates for each problem.
    \end{enumerate}
    \item \textbf{Blank fillers: role names}.
    \begin{enumerate}
        \item We randomly choose a role name \texttt{X}, and use \texttt{X's father}, \texttt{X's mother}, \texttt{X's brother}, \texttt{X's sister} and \texttt{X's neighbor} as the overlapped role names. 
        \item We choose from the name set \texttt{\{Ada, David, Emma, Jack, John, Mary, Max, Tom\}} for non-overlapped role names. 
        \item We write five names that have overlap with the original character, and five names that do not have overlap for each problem. 
    \end{enumerate}
    \item \textbf{Blank fillers: numbers}.
    \begin{enumerate}
        \item For in-range numbers, we randomly sample positive integers in the range of $[\frac{\ell}{10}, 10r]$, where $\ell$ and $r$ denote the smallest and the largest number that appear in the problem description and standard solution, respectively.
        \item For out-of-range numbers, we choose from the range of $[2, +\infty) \backslash [\frac{\ell}{10}, 10r]$.
        For very few problems that $\ell$ is relatively large (i.e., $\ell > 10^5$) where we choose out-of-range numbers from the range of $[2, \frac{\ell}{10}]$; for other problems we choose out-of-range numbers $n = a\times 10^b$ from the range $[10r, \infty)$, where $a$ and $b$ are both non-negative integers. 
        \item We write four in-range numbers and four out-of-range numbers for each problem. 
    \end{enumerate}
    \item Finally, if adding the irrelevant sentence causes ambiguity (e.g., \cref{tab:appendix:ambiguity}), we fix the question to ensure that the standard solution to the generated problem remain the same as the base problem. 
    \begin{table}[t]
    \centering
    \begin{tabular}{lp{0.65\textwidth}}
    \toprule
    \textbf{\textit{Original Problem}} & 
    Kim plants 80 cherry pits. 25\% of them sprout and Kim sells 6 of the saplings. How many cherry saplings does \underline{she} have left? \\
    \midrule 
    \textbf{\textit{Added Sentence}} & 
    Kim's mother plants 20 more potatoes. \\
    \midrule 
    \textbf{\textit{Ambiguous Problem}} & Kim plants 80 cherry pits. 25\% of them sprout and Kim sells 6 of the saplings. Kim's mother plants 20 more potatoes. How many cherry saplings does \underline{she} have left? \\\\
    \midrule 
    \textbf{\textit{Fixed Problem}} & Kim plants 80 cherry pits. 25\% of them sprout and Kim sells 6 of the saplings. Kim's mother plants 20 more potatoes. How many cherry saplings does Kim have left? \\
    \bottomrule
    \end{tabular}
    \caption{An example that adding irrelevant contexts causes ambiguity: after adding the sentence, it is unclear whether \underline{she} refers to Kim or Kim's mother. To ensure that the standard answer is unchanged, we modify the final question to make it clear and faithful to the original problem. }
    \label{tab:appendix:ambiguity}
\end{table}
\end{enumerate}

\section{Sample Predictions on \noisygsm}
\label{sec:appendix-sample-output}
In addition to the example outputs shown in \cref{fig:example-main}, we include more example problems and the predictions by different techniques (\cref{tab:example-full-1,tab:example-full-2}).

\begin{table*}[t]
    \centering \small
    \begin{tabular}{cp{0.45\textwidth}p{0.4\textwidth}}
        \toprule
        \bf Method 
         & \bf Original GSM8K Problem & \bf Modified Problem (w/ Irrelevant Sentence) \\
         & Jessica is six years older than Claire. In two years, Claire will be 20 years old. How old is Jessica now? & Jessica is six years older than Claire. In two years, Claire will be 20 years old. Twenty years ago, the age of Claire's father is 3 times of Jessica's age. How old is Jessica now? \\
         \midrule 
         \chainofthought & In two years, Claire will be 20 years old. So, Claire is 18 years old now. Jessica is six years older than Claire, so Jessica is 18 + 6 = 24 years old now. The answer is 24. & Jessica is six years older than Claire, so Jessica is 20 - 6 = 14 years old. Twenty years ago, the age of Claire's father is 3 times of Jessica's age, so 20 years ago, the age of Claire's father is 3 * 14 = 42 years old. The answer is 14. \\
         \midrule 
        \leasttomost & Let's break down this problem: 1. How old is Claire now? 2. How old is Jessica now? & A: Let's break down this problem: 1. How old is Claire now? 2. How old is Jessica now? \\
& 1. In two years, Claire will be 20 years old. So Claire is 20 - 2 = 18 years old now. & 1. In two years, Claire will be 20 years old. So Claire is 18 years old now. \\
& 2. Jessica is six years older than Claire. So Jessica is 18 + 6 = 24 years old now. & 2. Twenty years ago, the age of Claire's father is 3 times of Jessica's age. So 20 years ago, Claire's father is 3 * Jessica's age. So Claire's father is 3 * Jessica's age now. So Claire's father is 3 * Jessica's age - 20 years old now. So Claire's father is 3 * Jessica's age - 20 + 18 = 3 * Jessica's age - 2 years old now. So Claire's father is 3 * Jessica's age - 2 + 6 = 3 * Jessica's age + 4 years old now. So Claire's father is 3 * Jessica's age + 4 - 18 = 3 * Jessica's age - 14 years old now. So Claire's father is 3 * Jessica's age - 14 + 20 = 3 * Jessica's age + 6 years old now. So 3 * Jessica's age + 6 = 3 * Jessica's age - 2 + 6 = 3 * Jessica's age + 4. So 3 * Jessica's age + 6 = 3 * Jessica's age + 4. So 6 = 4. So Jessica's age = 2. So Jessica is 2 years old now.\\
& The answer is 24. & 
The answer is 2. \\
\midrule 
\program & Claire\_age = 20 - 2 & Claire\_age = 20 - 2 \\
& Jessica\_age = Claire\_age + 6 & Jessica\_age = Claire\_age - 6 \\
& print(Jessica\_age) & Claire\_father\_age = 3 * Jessica\_age \\
& & print(Jessica\_age) \\
        \bottomrule
    \end{tabular}
    \centering
    \caption{Output example \#1 for different prompting techniques, on a GSM8K problem and a corresponding \noisygsm problem.
    While all techniques give correct answers to the original problem, they output different different results for the modified \noisygsm problem.}
    \label{tab:example-full-1}
\end{table*}

\begin{table*}[t]
    \centering \small
    \begin{tabular}{cp{0.4\textwidth}p{0.4\textwidth}}
        \toprule
        \bf Method 
         & \bf Original GSM8K Problem & \bf Modified Problem (w/ Irrelevant Sentence) \\
         & Lucy has \$65 in the bank. She made a \$15 deposit and then followed by a \$4 withdrawal. What is Lucy's bank balance?  & Lucy has \$65 in the bank. She made a \$15 deposit and then followed by a \$4 withdrawal. Lucy's mother's monthly rent is \$10. What is Lucy's bank balance? \\
         \midrule 
         \chainofthought & Lucy has \$65 in the bank. She made a \$15 deposit, so she has \$65 + \$15 = \$80. She then made a \$4 withdrawal, so she has \$80 - \$4 = \$76. The answer is \$76. & Lucy has \$65 in the bank. She made a \$15 deposit and then followed by a \$4 withdrawal. Lucy's mother's monthly rent is \$10. Lucy's bank balance is \$65 + \$15 - \$4 - \$10 = \$66. The answer is \$66. \\
         \midrule 
        \leasttomost & Let's break down this problem: 1. How much money did Lucy deposit? 2. How much money did Lucy withdraw? 3. How much money does Lucy have in the bank? & Let's break down this problem: 1. How much money did Lucy deposit? 2. How much money did Lucy withdraw? 3. How much money does Lucy have in the bank? \\
& 1. Lucy deposited \$15. & 1. Lucy deposited \$15.\\
& 2. Lucy withdrew \$4. & 2. Lucy withdrew \$4. \\
& 3. Lucy has \$65 + \$15 - \$4 = \$76 in the bank. & 3. Lucy has \$65 + \$15 - \$4 - \$10 = \$66 in the bank.\\
& The answer is \$76. &  The answer is \$66. \\
\midrule 
\program & Lucy\_balance = 65 & Lucy\_bank = 65\\
& Lucy\_balance = Lucy\_balance + 15 & Lucy\_bank = Lucy\_bank + 15 \\
& Lucy\_balance = Lucy\_balance - 4 & Lucy\_bank = Lucy\_bank - 4 \\
& print(Lucy\_balance) & Lucy\_bank = Lucy\_bank - 10\\
&& print(Lucy\_bank) \\
        \bottomrule
    \end{tabular}
    \centering
    \caption{Output example \#2 for different prompting techniques, on a GSM8K problem and a corresponding \noisygsm problem.
    While all techniques give correct answers to the original problem, they output the same incorrect result for the modified \noisygsm problem.}
    \label{tab:example-full-2}
\end{table*}

\section{Full prompts in experiments}
\label{sec:appendix-prompt}
We list the prompts for all experiments in \cref{tab:all-prompts,tab:all-prompts-instructed}.
\begin{table}[t]
    \centering \small 
    \begin{tabular}{ccp{0.6\textwidth}}
         \toprule
         \bf Method & \bf Example & \multicolumn{1}{c}{\bf Prompt}\\ 
         & \text{w/ IrrCtx?} & \\
         \midrule         
         \chainofthought & \xmark & Q: Elsa has 5 apples. Anna has 2 more apples than Elsa. How many apples do they have together? \\
         & & A: Anna has 2 more apples than Elsa, so Anna has 2 + 5 = 7 apples. Elsa and Anna have 5 + 7 = 12 apples together. The answer is 12.\\
         & & Q: [Problem of Interest] \\
         & & A: \\
         \midrule 
         \chainofthought & \cmark & Q: Elsa has 5 apples. Anna has 2 more apples than Elsa. Liz has 4 peaches. How many apples do they have together? \\
         & & A: Anna has 2 more apples than Elsa, so Anna has 2 + 5 = 7 apples. Elsa and Anna have 5 + 7 = 12 apples together. The answer is 12.\\
         & & Q: [Problem of Interest] \\
         & & A: \\
         \midrule
         \leasttomost & \xmark & Q: Elsa has 5 apples. Anna has 2 more apples than Elsa. How many apples do they have together? \\
         && A: Let's break down this problem: 1. How many apples does Anna have? 2. How many apples do Elsa and Anna have together? \\
         && 1. Anna has 2 more apples than Elsa. So Anna has 2 + 5 = 7 apples. \\
         && 2. Elsa and Anna have 5 + 7 = 12 apples together. \\
         & & Q: [Problem of Interest]\\
         & & A: Let's break down this problem:\\
         \midrule
         \leasttomost & \cmark & Q: Elsa has 5 apples. Anna has 2 more apples than Elsa. Liz has 4 peaches. How many apples do they have together? \\
         && A: Let's break down this problem: 1. How many apples does Anna have? 2. How many apples do Elsa and Anna have together? \\
         && 1. Anna has 2 more apples than Elsa. So Anna has 2 + 5 = 7 apples. \\
         && 2. Elsa and Anna have 5 + 7 = 12 apples together. \\
         & & Q: [Problem of Interest]\\
         & & A: Let's break down this problem:\\
         \midrule
         0-\chainofthought & N/A & Q: [Problem of Interest] \\
         && A: Let's think step by step: \\
         \midrule 
         \program & \xmark & Q: Elsa has 5 apples. Anna has 2 more apples than Elsa. How many apples do they have together? \\
         && A: Let’s solve the problem by a Python program: \\
         && Elsa\_apples = 5 \\
         && Anna\_apples = 2 + Elsa\_apples \\
         && Elsa\_Anna\_apples = Elsa\_apples + Anna\_apples \\
         && print(Elsa\_Anna\_apples)\\
         && Q: [Problem of Interest] \\
         && A: Let’s solve the problem by a Python program: \\
         \midrule 
         \program & \cmark & Q: Elsa has 5 apples. Anna has 2 more apples than Elsa. Liz has 4 peaches. How many apples do they have together? \\
         && A: Let’s solve the problem by a Python program: \\
         && Elsa\_apples = 5 \\
         && Anna\_apples = 2 + Elsa\_apples \\
         && Elsa\_Anna\_apples = Elsa\_apples + Anna\_apples \\
         && print(Elsa\_Anna\_apples)\\
         && Q: [Problem of Interest] \\
         && A: Let’s solve the problem by a Python program: \\
         \bottomrule
    \end{tabular}
    \caption{Prompts used for all investigated techniques, without instruction. The placeholder [Problem of Interest] is substituted for each problem at the test time. }
    \label{tab:all-prompts}
\end{table}

\begin{table}[t]
    \centering \small 
    \begin{tabular}{ccp{0.75\textwidth}}
         \toprule
         \bf Method & \bf Example & \multicolumn{1}{c}{\bf Prompt}\\ 
         & \text{w/ IrrCtx?} & \\
         \midrule         
         \chainofthought & \xmark & Solve grade school math problems. Feel free to ignore irrelevant information given in the questions. \\
         && Q: Elsa has 5 apples. Anna has 2 more apples than Elsa. How many apples do they have together? \\
         & & A: Anna has 2 more apples than Elsa, so Anna has 2 + 5 = 7 apples. Elsa and Anna have 5 + 7 = 12 apples together. The answer is 12.\\
         & & Q: [Problem of Interest] \\
         & & A: \\
         \midrule 
         \chainofthought & \cmark & Solve grade school math problems. Feel free to ignore irrelevant information given in the questions. \\
         && Q: Elsa has 5 apples. Anna has 2 more apples than Elsa. Liz has 4 peaches. How many apples do they have together? \\
         & & A: Anna has 2 more apples than Elsa, so Anna has 2 + 5 = 7 apples. Elsa and Anna have 5 + 7 = 12 apples together. The answer is 12.\\
         & & Q: [Problem of Interest] \\
         & & A: \\
         \midrule
         \leasttomost & \xmark & Solve grade school math problems. Feel free to ignore irrelevant information given in the questions. \\
         && Q: Elsa has 5 apples. Anna has 2 more apples than Elsa. How many apples do they have together? \\
         && A: Let's break down this problem: 1. How many apples does Anna have? 2. How many apples do Elsa and Anna have together? \\
         && 1. Anna has 2 more apples than Elsa. So Anna has 2 + 5 = 7 apples. \\
         && 2. Elsa and Anna have 5 + 7 = 12 apples together. \\
         & & Q: [Problem of Interest]\\
         & & A: Let's break down this problem:\\
         \midrule
         \leasttomost & \cmark & Solve grade school math problems. Feel free to ignore irrelevant information given in the questions. \\
         && Q: Elsa has 5 apples. Anna has 2 more apples than Elsa. Liz has 4 peaches. How many apples do they have together? \\
         && A: Let's break down this problem: 1. How many apples does Anna have? 2. How many apples do Elsa and Anna have together? \\
         && 1. Anna has 2 more apples than Elsa. So Anna has 2 + 5 = 7 apples. \\
         && 2. Elsa and Anna have 5 + 7 = 12 apples together. \\
         & & Q: [Problem of Interest]\\
         & & A: Let's break down this problem:\\
         \midrule
         0-\chainofthought & N/A & Solve grade school math problems. Feel free to ignore irrelevant information given in the questions. \\
         &&  Q: [Problem of Interest] \\
         && A: Let's think step by step: \\
         \midrule 
         \program & \xmark & Solve grade school math problems. Feel free to ignore irrelevant information given in the questions. \\
         && Q: Elsa has 5 apples. Anna has 2 more apples than Elsa. How many apples do they have together? \\
         && A: Let’s solve the problem by a Python program: \\
         && Elsa\_apples = 5 \\
         && Anna\_apples = 2 + Elsa\_apples \\
         && Elsa\_Anna\_apples = Elsa\_apples + Anna\_apples \\
         && print(Elsa\_Anna\_apples)\\
         && Q: [Problem of Interest] \\
         && A: Let’s solve the problem by a Python program: \\
         \midrule 
         \program & \cmark & Solve grade school math problems. Feel free to ignore irrelevant information given in the questions. \\
         && Q: Elsa has 5 apples. Anna has 2 more apples than Elsa. Liz has 4 peaches. How many apples do they have together? \\
         && A: Let’s solve the problem by a Python program: \\
         && Elsa\_apples = 5 \\
         && Anna\_apples = 2 + Elsa\_apples \\
         && Elsa\_Anna\_apples = Elsa\_apples + Anna\_apples \\
         && print(Elsa\_Anna\_apples)\\
         && Q: [Problem of Interest] \\
         && A: Let’s solve the problem by a Python program: \\
         \bottomrule
    \end{tabular}
    \caption{All prompts with instructions. The placeholder [Problem of Interest] is substituted for each problem at the test time. }
    \label{tab:all-prompts-instructed}
\end{table}
%%%%%%%%%%%%%%%%%%%%%%%%%%%%%%%%%%%%%%%%%%%%%%%%%%%%%%%%%%%%%%%%%%%%%%%%%%%%%%%
%%%%%%%%%%%%%%%%%%%%%%%%%%%%%%%%%%%%%%%%%%%%%%%%%%%%%%%%%%%%%%%%%%%%%%%%%%%%%%%

\end{document}